# Improvement of image classification by multiple optical scattering

Xinyu Gao, Yi Li*, Yanqing Qiu, Bangning Mao, Miaogen Chen, Yanlong Meng, Chunliu Zhao, Juan Kang, Yong Guo, and Changyu Shen

*Abstract*—Multiple optical scattering occurs when light propagates in a non-uniform medium. During the multiple scattering, images were distorted and the spatial information they carried became scrambled. However, the image information is not lost but presents in the form of speckle patterns (SPs). In this study, we built up an optical random scattering system based on an LCD and an RGB laser source. We found that the image classification can be improved by the help of random scattering which is considered as a feedforward neural network to extracts features from image. Along with the ridge classification deployed on computer, we achieved excellent classification accuracy higher than 94%, for a variety of data sets covering medical, agricultural, environmental protection and other fields. In addition, the proposed optical scattering system has the advantages of high speed, low power consumption, and miniaturization, which is suitable for deploying in edge computing applications.

## I. INTRODUCTION

Machine learning plays an essential role in many fields, such as image classification, image segmentation, semantic recognition, image enhancement and so on. At the same time, machine learning has a huge demand for computing resource, which brings a tremendous burden to computing devices. In this context, optical computing becomes clearly promising because of fast calculation and low power consumption. Thanks to the high degree-of-freedom of light (wavelength, polarization, mode, phase, intensity, etc.), optical computing can handle parallel tasks quickly and the application of passive optical component brings the advantage of low power consumption. Recently, many efforts have been made in optically accelerated machine learning computations, such as multilayer perceptrons, convolutional neural networks [1, 2], spiking neural networks [3-5] and reservoir computing [6-11].

However, many expensive optical active controlling devices such as optical comb, SLM, or high bandwidth Electro-optic modulator are required in the above studies. They are hard to apply in the field of edge computing due to the complex structure and high cost. Saade et al. reported an optical system using random scattering media and digital micromirror devices (DMD) with compact size and low cost [12]. DMD reflected the image onto a random scattering medium, thus performing a random projection on the data. It has been proved that the random projection can reduce the dimensions of the data while preserving the distance between the data as much as possible [13, 14]. The data after dimension reduction can be classified more efficiently [15]. However, limited by the working principle, the DMD can only present two states, 0 or 1, and image binarization was required before sending to the system. Nevertheless, binary encoding sacrifices space occupation to express the RGB information. Moreover, the binary pattern intuitively changes the original pattern features, and its influence on the classification effect is not clear yet.

In this study, we developed an optical random scattering system that fully utilizes the degrees of freedom of light. By using LCD screen instead of DMD, the image processing is

National Key R&D Program of China (Grant 2019YFE0112000, Grant 2020YFF0217803), Zhejiang Xinmiao Talents Program (Grant 2021R409043). National Natural Science Foundation of China (Grant 11874332, Grant 61727816, Grant 61775202), Key R & D plan of Zhejiang Province (Grant 2021C01179), Zhejiang Provincial Natural Science Foundation of China (Grant LY21F050006).

Xinyu Gao, Yi Li, Yanqing Qiu, Bangning Mao, Yanlong Meng, Chunliu Zhao, Juan Kang, Changyu Shen are with the China Jiliang University, Hangzhou 310018 CHN (e-mail: yil@cjlu.edu.cn).
Miaogen Chen Key Laboratory of Intelligent Manufacturing Quality Big Data Tracing and Analysis of Zhejiang Province, Department of Physics, China Jiliang University, Hangzhou 310018, CHN.
Yong Guo is with the Zhejiang Zhongnong Engineering Testing Co. Ltd, Hangzhou 310018 CHN.

upgraded from binary data to three-channel grayscale data. This saves the encoding process and brings a higher degree-of-freedom for classification with lower cost. Our random scattering system combined with the ridge classification (RC) to form an extreme learning machine (ELM) [16], which can efficiently classifies the data. In the classification tasks of multiple data sets, scattering scheme has achieved a classification accuracy of more than 94%, presenting a significant improvement compared to the pure ridge classification on the original images.

## II. EXPERIMENT

The implementation of the proposed random scattering system is illustrated in Fig.1. The light source is a white light laser composed of RGB three laser diodes with the output wavelength of 445nm, 520nm and 638nm. For the LCD screen, its backlight illumination part has been removed, and only liquid crystal layer, polarizers and filter layer were left. The images were sent to the LCD via a computer. Thanks to the RGB light source and the LCD filter layer, the image can be displayed at different wavelengths. And subsequently were scattered by a random scattering medium (silica slide covered with a thin transparent matte paper). Lights' propagation direction, phase, and amplitude were randomly modified. An RGB camera (Mindvision mv-sua500c-t) with the resolution of 2952×1944 was used to record the SPs.

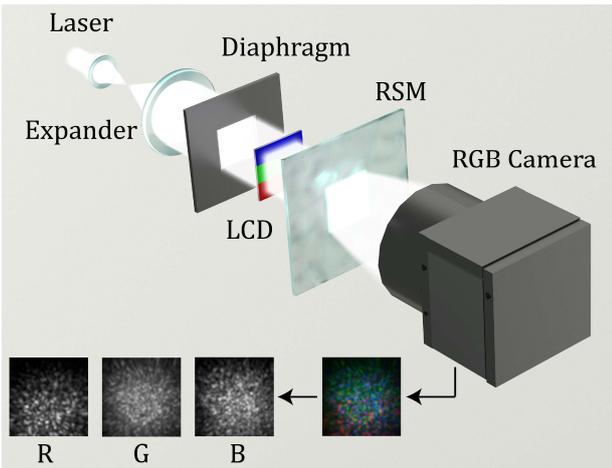

Fig. 1. Structure diagram of experimental setup. The SP also has 3 channels and the pattern of each channel is unique, as shown at the bottom of the figure.

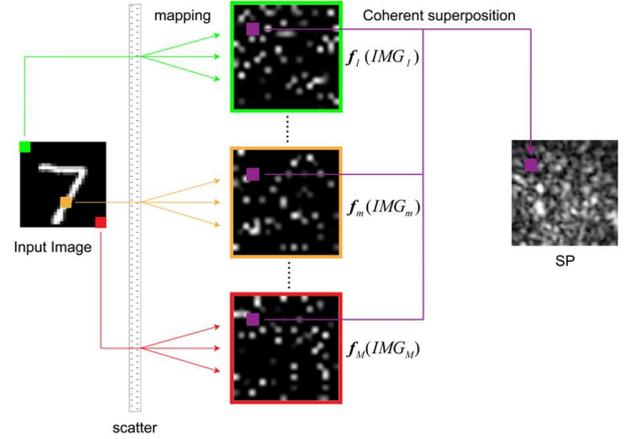

Fig.2. Schematic diagram of random scattering and the generated speckles.

The process of scattering randomly changes the amplitude, phase, and direction of the input light so that the original image is scrambled. Our simulations show that the scrambled light from single pixel may spread to a quite large area. Light scrambling can be regarded as a linear mapping. Each pixel is mapped into a light field by the point spread function $f_m$ that is determined by the pixel position and the random scattering medium. Similar to the studies of Zhou et al. and Lin et al.[17, 18], we can regard random scattering as single-hidden layer feedforward neural network that can extract the features of input images. Suppose that input image is of M pixels, and $IMG_m$ represents the m-th pixel. Then the light field after random scattering can be denoted as $f_m(\mathbf{IMG}_m)$. As shown in the Fig.2, the SP was generated by the coherent superposition of $f_m(\mathbf{IMG}_m)$. Suppose that SP has N pixels, and the n-th pixel can be expressed as Eq.1, where g is a nonlinear activation function representing the coherent superposition. That is to say, each input image pixel undergoing linear transformation has an impact on the value of the $SP_n$.

$$\mathbf{SP}_n = g_n[f_1(\mathbf{IMG}_1)_n, f_2(\mathbf{IMG}_2)_n, \cdots f_M(\mathbf{IMG}_M)_n] \quad (1)$$
$$n = 1, 2, \cdots N$$

In this study, RC was used for image classifications, which is an improved least squares estimation method and is quite suitable for the edge computing. The process of classification according to SP is shown in equation 2, where c is a classifier, L the label, and β the weight of the hidden layer to the output calculated by ridge regression. Thus, the random scattering process shown in equation 1 and 2 can be considered as the classification by ELM with N hidden nodes [16, 19-21].

$$L = c(\boldsymbol{\beta} \cdot \mathbf{SP}) \quad (2)$$

ELM is a kind of machine learning system based on feedforward neuron network [22]. It has the characteristics of fast training speed, strong generalization ability, and strong classification ability [21]. The approximation ability of the feedforward neuron network is proportional to the number of hidden nodes. However, in traditional computing systems, the increase of hidden nodes will consume more computing resources. Based on the above analysis, the optical methods could increase the hidden nodes in a cost-effective way, thereby improving the classification ability of the algorithm.

III. RESULT

Nine data sets were selected to evaluate the performance of the optical random scattering system. Different from the previous studies [12], all the input images were pre-processed into RGB 3-channel pictures with the spatial resolution of 48×48. In this way, the degree-of-freedom brought by the wavelength can be fully utilized. The pre-processed images were then sent to the LCD screen and illuminated by the RGB laser. All the SPs were recorded by the camera. It is noted that the resolution of SP was 80×80. In the following RC classification, the SPs were reshaped into one-dimensional arrays with a length of 19200 (80×80×RGB), implying the hidden nodes of ELM as high as 19200.

Take the HAM10000 data set as an example, which is used to diagnose pigmented skin lesions with 7 categories of data [23]. Figure 3 shows the typical images of 7 types of images in the HAM10000 dataset and their corresponding pre-processed images and SPs. In order to ensure the repeatability of the classification results, all data are randomly split into training and test set, accounted for 85% and 15% of the total data, respectively.

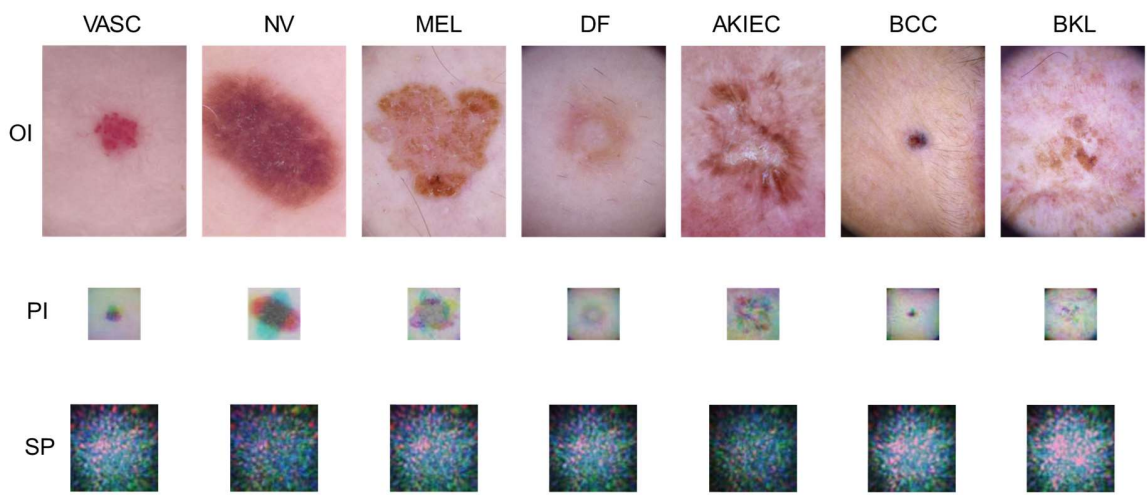

Fig. 3. Typical pictures of HAM10000 dataset. This data set contains 7 types of lesion images, including BKL, BCC, AKIEC, DF, MEL, NV and VASC. OI represents the original image, PI is the pre-processed image.

In addition to HAM10000, the remaining 8 data sets include the commonly used MNIST, Chest_X-ray (lung X-ray images [24]), Malaria (cell microscopy taken with a smartphone and an optical microscope [25]), Leukemia (microscopic cells [26]), Lung&Colon (Lung and Colon Cancer Histopathological Images [27]), Tomato(from PlantVillage-Dataset [28]), and Garbage (garbage classification [29]).

Fig.4 (a) and (b) summarize the test results of MNIST and Fashion-MNIST in the form of confusion matrix.

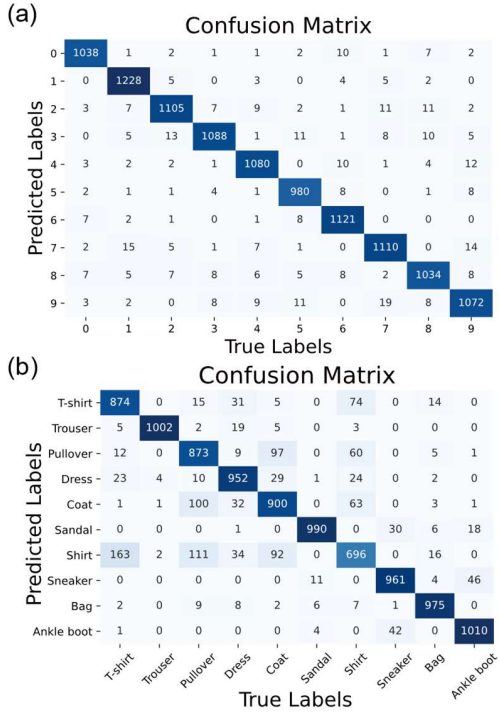

Fig. 4. Classification results by using our system. (a) Confusion matrix of MNIST data sets, and (b) Confusion matrix of Fashion-MNIST data sets.

As shown in Table 1, the classification accuracy by RC, ResNet-18 [30], with and without random scattering system of were compared together. All the algorithms used in the experiment were deployed with Python. The ResNet-18 was implemented in the Pytorch framework on a GeForce RTX 3080 GPU. The batch size of model was 50 and the optimizer was the SGD with momentum = 0.9 and weight_decay = $10^{-3}$. The learning rate (LR) was initialized as $10^{-3}$. Training was stopped at the 50th epochs, and the accuracy of the ResNet-18 is the highest one in the validation set.

TABLE I
COMPARISON OF THE ACCURACY AMONG THE THREE METHODS

| Datasets | Chest_X-ray | MNIST | Lung&Colon |
|---|---|---|---|
| **RSS + RC** | 0.999 | 0.966 | 0.998 |
| RC | 0.899 | 0.851 | 0.614 |
| ResNet-18 | 0.948 | 0.999 | 0.931 |
| SC+RC | 0.997 | 0.950 | 0.996 |
| Datasets | Leukemia | Tomato | Fashion-MNIST |
| **RSS + RC** | 0.946 | 0.990 | 0.885 |
| RC | 0.692 | 0.531 | 0.820 |
| ResNet-18 | 0.800 | 0.644 | 0.957 |
| SC+RC | 0.939 | 0.980 | 0.835 |

| Datasets | HAM10000 | Malaria | Garbage |
|---|---|---|---|
| **RSS + RC** | 0.974 | 0.993 | 0.962 |
| RC | 0.640 | 0.635 | 0.375 |
| ResNet-18 | 0.704 | 0.878 | 0.550 |
| SC+RC | 0.964 | 0.988 | 0.936 |

Note: RSS + RC represents the random scattering scheme using RGB light sources. SC+RC represents the random scattering scheme using a single wavelength light source (638nm). RC and ResNet-18 represent classification of the original data without random scattering. All the results are the averaged values of 5 tests.

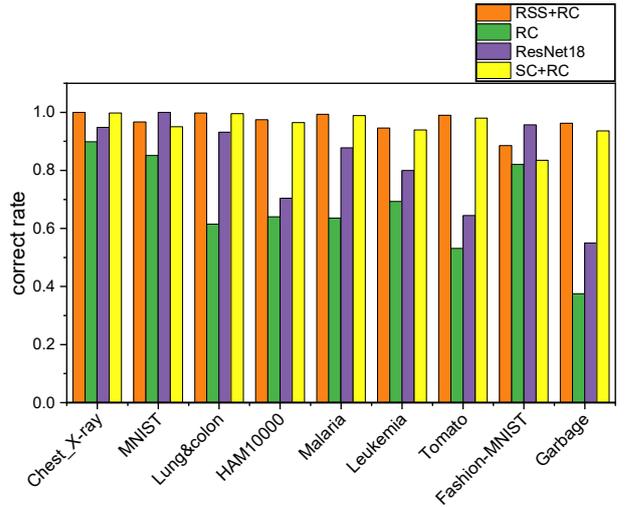

Fig.5. Comparison of correct rate of three schemes.

For more intuitive comparison, we established a histogram as shown in Fig. 5. For 8 data sets except Fashion-MNIST, the test accuracy rates are above 94%. Obviously, the comprehensive classification effect of random scattering system + RC is the best among the four schemes, only inferior to the CNN method in individual data sets. It turns out that the images data after scattering can be classified more accurately. And the accuracy is significantly higher than that of the non-scattering scheme, regardless of whether the classification algorithm is RC or ResNet-18. When random scattering system has only a single channel, there are only 1/3 hidden units of ELM compared to the three-channel scheme. Therefore, for some complex data sets, the 3-channel scheme has obvious advantages because of larger number of hidden units.

Moreover, although the classification ability of the CNN algorithm surpasses the random scattering method on individual data sets, the training process took a longer time. The time-consuming comparison with RC is shown in Table.

2. The training of ResNet-18 was completed on the RTX3080 GPU which possesses 8704 Compute Unified Device Architecture (CUDA) cores. In contrast, RC runs on the Xeon W-2140B with only 8 cores. However, although RC uses far fewer computing resources than ResNet-18, its calculation speed is still much faster than ResNet-18. It is worth noting that the time consumed was recorded of ResNet-18 when it obtains the best classification result.

TABLE II
COMPARISON OF COMPUTING TIME BETWEEN RESNET-18 AND RC

| Datasets | Chest_X-ray | MNIST | Lung&Colon |
|---|---|---|---|
| ResNet18 | 125.46s | 482.73s | 481.70s |
| RC | 1.35s | 41.82s | 20.94s |
| Datasets | Leukemia | Tomato | Fashion-MNIST |
| ResNet18 | 331.21s | 470.21s | 1092.12s |
| RC | 4.25s | 20.18s | 39.52s |
| Datasets | HAM10000 | Malaria | Garbage |
| ResNet18 | 214.69s | 769.35s | 267.39s |
| RC | 3.33s | 21.87s | 6.42s |

Obviously, the calculation of RC is much faster than that of ResNet-18, with lower demand of computing resources. The random scattering made a large number of random feature extractions of the image and allowing the algorithm to approximate samples with a slight error. Thanks to the parallel processing capabilities of optics, random scattering feature extractions can be quickly implemented with lower computational burden.

IV. CONCLUSION

In this study, we used the random scattering to improve the classification accuracy for RC, taking advantage of the high parallelism and high degree-of-freedom of light. Random scattering system and RC together form an ELM. Thanks to the passive random scattering medium, ELM contains a large number of hidden units while maintaining low power consumption and fast convergence. Therefore, the accuracy of classification can be improved when processing nonlinear data classification. At the same time, the characteristics of RC that fast calculation and low demand for computing resources are retained. In addition, this optical random scattering system can be miniaturized because the used components (laser, LCD, RGB camera, etc.) have compact size alternatives.

The proposed random scattering system scheme should have broad application prospects in the field of lower power edge computing, such as preliminary screening of skin diseases, sorting trash types, or identifying the types of pests and diseases.

Although light propagation in our optical setup provides fully parallel features extraction, the parallelism of random scattering system still has space for further improvement, such as using space-division multiplexing to scatter multiple images.